# A many-objective evolutionary algorithm using indicator-driven weight vector optimization

Xiaojing Han*
School of Mathematics and Statistics
Qinghai Normal University
Xining,810016, China
Hanxiaojing@163.com

Yuanxin Li
School of Mathematics and Statistics
Qinghai Normal University
Xining,810016, China
545378560@qq.com

*Abstract*—For regular Pareto Fronts (PFs), such as those that are smooth, continuous, and uniformly distributed, using fixed weight vectors is sufficient for multi-objective optimization approaches using decomposition. However, when encountering irregular PFs—including degenerate, disconnected, inverted, etc. Fixed weight vectors can often cause a non-uniform distribution of the sets or even poor optimization results. To address this issue, this study proposes an adaptive many-objective evolutionary algorithm with a simplified hypervolume indicator. It synthesizes indicator assessment techniques with decomposition-based methods to facilitate self-adaptive and dynamic adjustment of the weight vectors in many-objective optimization methods. Specifically, based on the MOEA/D framework, it uses a simplified hypervolume indicator to accurately assess solution distribution. Simultaneously, applying the R2 indicator (as an approximation of hypervolume) dynamically regulates the update frequency of the weight vectors. Experimental results demonstrate that the proposed algorithm is efficient and effective when compared with six state-of-the-art algorithms.

Keywords- Many-Objective Evolutionary Algorithm; Pareto Fronts; Weight Adjustment; Hypervolume

## I. INTRODUCTION

Real-world optimization problems across various domains, such as engineering design[1], biotechnology[2] often involve two or more conflicting objectives, making it difficult to achieve a balanced outcome during the optimization process. Such problems are called multi-objective optimization problems (MOPs). When the number $m$ of optimization objectives exceeds three, it is called a many-objective optimization problem (MaOPs). Without loss of generality, an MOP/MaOP can be defined as

$$\min F(x) = (f_1(x), f_2(x), \dots f_m(x))^T$$
$$s.t. \quad x \epsilon \Omega \qquad (1)$$

To enhance the performance of most MOEAs in solving MaOPs, many methods have been proposed, which can be roughly classified into three categories.

The first category comprises MaOEAs based on Pareto dominance relations. However, as the number of objectives increases, this type of algorithm typically experiences a decrease in the effectiveness of deterministic dominance relations, which results in a reduction in the selection pressure exerted by Pareto dominance. To address this issue, alternative strategies such as gws-dominance [3] and PRV-dominance[4] have been proposed to enhance the selection pressure on PF by modifying the dominance relations.

The second category comprises MaOEAs based on performance indicators. This type of method employs metrics such as convergence and distribution to identify Pareto optimal solutions. Many scholars commonly use performance indicators such as Hypervolume(HV)[5] and Inverted Generational Distance (IGD)[6] to measure the quality of individuals. In recent years, *R2*-based MaOEAs have gained increasing popularity due to comprehensive properties, which evaluate exactly the quality of the solution sets[7].

The third category comprises decomposition-based MaOEAs. This method decomposes multi-objective problems into multiple single-objective optimization-related sub-problems and employs heuristic search to collaboratively optimize these sub-problems. However, MOEA/D largely depends on the shape of the Pareto front (PF) of the problem[8]. A direct and effective strategy is to dynamically adjust the weight distribution during the evolutionary process. At present, several weight adaptation methods have been proposed, in which the sparsity of individuals is evaluated based on the distance to their k-th nearest neighbor to determine where to add or remove weights. Meanwhile, during the population evolution, the timing and frequency of weight vector updates play a critical role in its adaptive adjustment.

Taking comprehensively all of the aforementioned factors into account, this study proposes a new indicator-driven many-objective evolutionary algorithm (IMOEA). The core of the IMOEA is the integration of indicator evaluation techniques with the decomposition method. Specifically, it operates within the MOEA/D framework and employs a simplified hypervolume indicator to assess individual sparsity. Meanwhile, it incorporates an approximate hypervolume method (R2) to dynamically control the frequency of weight vector updates. Finally, IMOEA effectively guides the search by the evolution of the weight vector, providing an efficient and precise solution set for many-objective optimization problems. The main contributions of this study are summarized as follows:

1) Using the simplified hypervolume to assess individual sparsity effectively reduces parameter uncertainty, thereby enabling precise updates of weights in critical regions.

2) By using approximate hypervolume method, the update frequency of the weight vectors is dynamically controlled, thereby mitigating the issue of excessively frequent or untimely adjustments.

3) The proposed algorithm demonstrates superior performance compared to existing algorithms on both regular and irregular test instances.

## II. RELATED WORK

### A. The method of decomposition

MOEA/D adopts three widely used decomposition methods, namely the weighted sum method[9], the Tchebycheff decomposition method (TCH), and the penalty-based boundary intersection method (PBI). This study employs the Tchebycheff method to guide the population evolution, and the mathematical formulation is presented as follows:

$$g^{tch}(x|\lambda, z^*) = \max_{1 \leq i \leq m}\{\lambda_i |f_i(x) - z_i^*|\} \quad (2)$$

Where, $\lambda$ represents the weight vector, satisfying $\sum_{i=1}^{m}\lambda_i = 1$, $z_i^* = min\{f_i(x)|x \epsilon \Omega,\}$, $i \epsilon \{1,2,...m\}$

### B. The Methods for generating weight vectors

There are mainly four methods for generating weight vectors. The first method is the simplex lattice design method proposed by Das and Dennis (DD)[10], which uses the weight vector to determine the population size, i.e., $N = C_{H+m-1}^{m-1}$. As the number of objective dimensions increases, the population size also increases. The second method is the double-layer weight generation method[11], i.e., the explosive growth of the number of weight vectors and the relative insufficiency of boundary points. The third method is the random sampling to generate the weight vector (UR)[12]. which allows for flexible population sizes but results in less predictable distribution quality. The fourth method is the weight generation method based on the TCH function proposed by Li et al.[13], which demonstrates superior performance in achieving uniform distribution, particularly for non-convex Pareto fronts (PFs).

## III. PROPOSED ALGORITHM

### A. Algorithm framework

Algorithm 1 shows the main process of IMOEA. First, generate a random population and select parents in neighborhoods with different probabilities. After crossover and mutation, compare offspring with parents via scalar functions. Replace parents if offspring are better. Meanwhile, maintain the external archive and adjust weights until some criterion is satisfied.

### B. Maintenance of external Archive

In IMOEA, an external archive is predefined to store non-dominated solutions generated during evolution. In this study, a niche-based technique is employed, individuals with higher crowding degrees are iteratively eliminated to maintain a representative subset of non-dominated solutions[13]. The crowding distance of an individual $p$ in the external archive is formulated as follows:

$$D(p) = 1 - \prod_{q \in A, q \neq p} R(p,q) \quad (3)$$

$$R(p,q) = \begin{cases} \frac{d(p,q)}{r}, & if\ d(p,q) \leq r \\ 1, & otherwise \end{cases} \quad (4)$$

Where, $d(p,q)$ denotes the Euclidean distance between individuals $p$ and $q$, The niche radius $r$ is set to the median of distances from all solutions to their $k$-th nearest neighbors. The value of $k$ is determined by the number of objectives. All objectives are normalized before calculations to ensure algorithm accuracy and comparability.

| Algorithm 1: IMOEA |
|---|
| **Input**: N(population size, the size of weights), T(size of neighborhood), $T_{max}$(the maximal generation), NA(the size of external archive EA), A(Mating pool) |
| **Output:** EA |
| 1    Initialize population (P={$x_i$}) , weight vector (W), Ideal point ($Z^*$) and Neighborhoods ($B_i$), EA ← Nondominanted solution set, gen=0 |
| 2    **while** gen< $T_{max}$ **do** |
| 3      **for** $i$=1:N **do** |
| 4        **if** rand(0,1)<δ,**then** A←($B_i$) |
| 5        **else** A←P //randomly chose |
| 6        **end if** |
| 7      $x' = $ OperatorGA($x^a, x^b$)// select two parents($x^a, x^b$) from mating pool A to generate offspring $x'$ |
| 8        Updating ideal point($Z^* = $ min ($Z^*, F(x')$) |
| 9        **if** $g^{tch}(x'|W,Z^*) \leq g^{tch}(x_i|W,Z^*)$ **then** $x_i \leftarrow x'$ //replacement |
| 10       **end if** |
| 11       Update EA |
| 12      **end for** |
| 13    **if** |NA|>2|N| |
| 14    Maintain EA with size of 2N //see III-B |
| 15    **end if** |
| 16    **if** 0.2* $T_{max}$≤gen≤0.9*$T_{max}$ |
| 17      **if** improvement-rate<threshold//see III-D |
| 18      Weight Adjustment //see Algorithm 2 |
| 19      **end if** |
| 20    **end if** |
| 21    gen=gen+1 |
| 22  **end while** |
| 23  **return EA** |

### C. Weight adjustment

In multi-objective optimization, the $k$-nearest neighbors method is commonly used to measure individual sparsity levels[14]. However, determining an appropriate $k$ value is challenging, many studies take the number of objectives as $k$. Inspired by those approaches, IMOEA introduces a simplified hypervolume-based strategy to evaluate population sparsity, in which the reference points are determined as the intersection formed by the maximum objective values among each individual's neighbors across all dimensions, and the hypervolume bounded by the individual and this reference point is then computed. As illustrated in Fig.1, a 2-D coordinate the sparsity level of individual A is represented by the shaded area. Solution C outperforms A in the $f_2$, and B outperforms A in $f_1$. The reference point is determined by distances $B'$ and $C'$. The calculation formula is as follows:

$$V_i = \prod_{j=1}^{M}(f_j(r_i) - f_j(x_i)) \quad (5)$$

Where, $x_i$ represents an individual in the population, $r_i$ and $V_i$ respectively denote the reference point and volume value of the i-th individual.

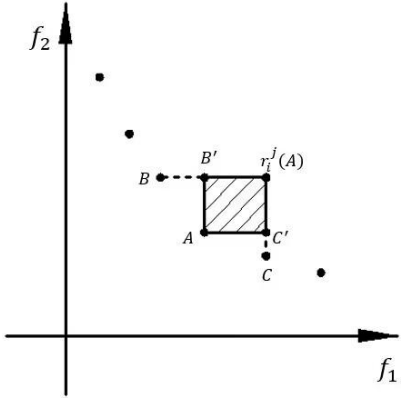

Fig.1.The simplified hypervolume represents the sparsity level of an individual.

Based on the simplified hypervolume calculation, a larger hypervolume indicates that the local region is relatively sparse. Therefore, some weights can be added in these corresponding sparse regions. The WS-transformation maps the weight vector of the scalar sub-problem to its solution mapping vector and is reflexive[14]. The method is as follows:

$$w = \left( \frac{\frac{1}{f_1 - Z_1^*}}{\sum_i^m \frac{1}{f_i - Z_i^*}}, \frac{\frac{1}{f_2 - Z_2^*}}{\sum_i^m \frac{1}{f_i - Z_i^*}}, \ldots, \frac{\frac{1}{f_m - Z_m^*}}{\sum_i^m \frac{1}{f_i - Z_i^*}} \right) \quad (6)$$

The reduction of weight vectors is determined by the magnitude of the individual hypervolume. We iteratively remove densely distributed weights until the predetermined threshold value is reached, which is the same as that in[14]. The pseudocode of weight adjustment is described in Algorithm 2.

**Algorithm 2**: Weight Adjustment
**Input**: W(Weight), EA(External Archive), P(Population), nus(predetermined threshold value), adjust←0
**Output**: P, W, $B_i$
1  **if** adjust<nus
2  Caculate sparsity level of each individual in population P by Eq. (5) to remove the highest sparsity level individual in P and associated weight vector //delete weight vector
3  adjust← $adjust + 1$
4  **end if**
5  **if** adjust>0
6  Caculate sparsity level of each individual in population P by Eq. (5) and (6) to add an individual in the lowest sparsity region and associated weight vector //add weight vector
7  adjust← $adjust - 1$, Updating neighborhood $B_i$
8  **end if**
9  **Return P, W, $B_i$**

*D. The frequency of adaptive update of the weight vector*

During population evolution, the update time and frequency of the weight vector are crucial for its adaptive adjustment. This is because changing the weight vector alters the sub-problems being optimized. If the vector is updated too frequently, the computational cost may be increased. Conversely, if it is not adjusted at all, its distribution may be significantly worse.

This study adopts the approximate hypervolume method proposed by Shang et al.[15], known as the $R_2^{new}$ indicator, which uses different line segments to approximate the hypervolume. Compared with traditional R2-based hypervolume approximation methods[16], it demonstrates significantly better computational efficiency than conventional approaches. The specific calculation formula is as follows:

$$R_2^{new}(s, EA, W) = \frac{1}{|W|} \sum_{\lambda \in W} L(s, EA, r, \lambda)^\alpha$$

$$= \frac{1}{|W|} \sum_{\lambda \in W} \max_{s \in EA} \{g^{mtch}(s|\lambda, r)\}^\alpha \quad (7)$$

$$g^{mtch}(s|\lambda, r) = \min_{j \in \{1,\ldots,m\}} \left\{ \frac{|s_j - r_j|}{\lambda_j} \right\} \quad (8)$$

Where, suggesting α = m indicates the approximation using the average m power length of the line segments, λ represents the direction vector, |W| indicates the cardinality of the direction vector, $r$ represents the reference point, and $s$ represents the solution.

Therefore, IMOEA uses the rate of change of $R_2^{new}$ to monitor whether adjustment to the weight vector is necessary. Moreover, as highlighted in [13], when populations approach the final stage of the optimization process, changes to the weight vector may cause the solution to evolve insufficiently along the specified search direction. Thus, IMOEA stops adjusting weight vector during the last 10% of the evaluation period. Meanwhile, once the $Improvement_{rate}$ exceeds the preset threshold (TH), the weight adjustment is implemented. The setting of threshold is provided in IV-C.

$$Improvement_{rate} = \frac{currentR_2^{new} - lastR_2^{new}}{lastR_2^{new}} \quad (9)$$

Where, $currentR_2^{new}$ and $lastR_2^{new}$ respectively denote $R_2^{new}$ value from current generation and previous generation.

## IV. EXPERIMENTAL RESULTS

This section presents the test results of IMOEA and six state-of-the-art algorithms on regular and irregular MaOPs, namely, ANSGA-III[17], Adaw[13], RVEA*[18], MOEA/D-AWA[14], MOEA/D-DU[19] and NRV-MOEA[20]. Specifically speaking, Adaw, MOEA/D-AWA, MOEA/D-DU are carried out under the decomposition framework and the weight vector is adjusted regularly, while ANSGA-III, RVEA*, and NRV-MOEA adjust the existing reference vector each time in the environmental se-lection. Four suits of test problems are consider for testing the compared algorithm, i.e., DTLZ benchmarks[21], IDTLZ benchmarks[22], WFG benchmarks[23], MaF benchmarks[24]. All algorithms are implemented on the PlatEMO platform [25] with MATLAB 2023a on Intel Core i5-8400 CPU@2.8GHZ.

Table1.HV values of solution sets obtained by seven algorithms on regular and irregular test instances

| Problem | M | D | ANSGAIII | AdaW | MOEADAWA | MOEADDU | RVEA | NRVMOEA | IMOEA |
|---|---|---|---|---|---|---|---|---|---|
| DTLZ1 | 5 | 9 | 9.0698e-1 (1.02e-1) - | 9.3863e-1 (1.21e-1) = | 9.5883e-1 (7.75e-3) = | 9.5027e-1 (7.78e-2) - | 9.6456e-1 (1.37e-2) + | 9.3171e-1 (1.10e-1) - | 9.6202e-1 (4.13e-3) |
|  | 10 | 14 | 5.4232e-1 (4.26e-1) - | 5.2824e-2 (1.49e-1) - | 8.7007e-1 (2.34e-2) - | 9.0793e-1 (1.74e-1) - | 9.4937e-1 (1.76e-1) - | 0.0000e+0 (0.00e+0) - | 9.8008e-1 (3.48e-2) |
|  | 15 | 19 | 6.2802e-1 (3.63e-1) - | 0.0000e+0 (0.00e+0) - | 9.4201e-1 (2.83e-1) + | 4.5561e-1 (3.53e-1) - | 9.4902e-1 (1.00e-1) + | 0.0000e+0 (0.00e+0) - | 9.0572e-1 (1.66e-1) |
| DTLZ3 | 5 | 14 | 3.5409e-2 (1.37e-1) - | 0.0000e+0 (0.00e+0) - | 4.3149e-1 (3.16e-1) - | 1.2175e-1 (2.17e-1) - | 1.6016e-1 (2.44e-1) - | 5.2423e-2 (1.46e-1) - | 5.9657e-1 (2.47e-1) |
|  | 10 | 19 | 0.0000e+0 (0.00e+0) - | 0.0000e+0 (0.00e+0) - | 2.3448e-1 (2.35e-1) = | 3.7786e-1 (1.44e-1) - | 4.4796e-2 (1.63e-1) - | 0.0000e+0 (0.00e+0) - | 3.2224e-1 (3.60e-1) |
|  | 15 | 24 | 0.0000e+0 (0.00e+0) - | 0.0000e+0 (0.00e+0) - | 5.4131e-1 (1.23e-1) + | 0.0000e+0 (0.00e+0) - | 3.7708e-2 (1.70e-1) - | 0.0000e+0 (0.00e+0) - | 1.5361e-1 (2.38e-1) |
| DTLZ5 | 5 | 14 | 9.2630e-2 (7.01e-3) - | 8.0108e-2 (2.04e-2) - | 1.1914e-1 (3.28e-4) - | 1.0489e-1 (1.35e-2) - | 9.1317e-2 (1.72e-3) - | 9.9774e-2 (1.11e-2) - | 1.2667e-1 (1.00e-3) |
|  | 10 | 19 | 5.4342e-2 (2.10e-2) - | 1.2426e-2 (1.86e-2) - | 9.6673e-2 (2.97e-4) - | 9.5007e-2 (5.94e-4) - | 9.1752e-2 (1.80e-3) - | 5.5502e-2 (1.86e-2) - | 1.0047e-1 (3.15e-4) |
|  | 15 | 24 | 5.0826e-2 (2.20e-2) - | 2.6973e-2 (2.40e-2) - | 9.2253e-2 (6.24e-4) - | 9.0421e-2 (8.31e-4) - | 9.1098e-2 (3.89e-4) - | 2.7113e-2 (2.34e-2) - | 9.5145e-2 (2.19e-4) |
| DTLZ7 | 5 | 24 | 2.3377e-1 (4.35e-3) + | 2.4250e-1 (3.34e-3) + | 9.2511e-2 (3.34e-2) - | 0.0000e+0 (0.00e+0) - | 2.0750e-1 (5.72e-3) + | 2.6117e-1 (4.89e-3) + | 1.4142e-1 (1.13e-2) |
|  | 10 | 29 | 9.0063e-2 (3.30e-2) + | 4.1133e-3 (5.29e-3) - | 1.2827e-2 (1.52e-2) - | 0.0000e+0 (0.00e+0) - | 7.7202e-2 (3.76e-2) + | 1.0138e-1 (1.33e-2) + | 4.7675e-2 (2.10e-2) |
|  | 15 | 34 | 1.3565e-1 (1.23e-2) + | 4.3856e-8 (2.40e-7) - | 2.4160e-3 (6.72e-3) = | 0.0000e+0 (0.00e+0) - | 6.7935e-2 (2.35e-2) + | 1.2212e-2 (8.34e-3) + | 4.1368e-3 (6.71e-3) |
| IDTLZ2 | 5 | 14 | 7.5617e-2 (5.88e-3) + | 9.5561e-2 (1.80e-3) + | 6.9623e-2 (2.05e-3) + | 4.5479e-3 (8.50e-3) - | 7.7641e-2 (6.00e-3) + | 1.1829e-1 (2.49e-3) + | 6.5323e-2 (2.81e-3) |
|  | 10 | 19 | 1.7270e-4 (9.86e-6) + | 2.2079e-5 (6.98e-6) - | 3.0282e-6 (4.39e-6) - | 2.7388e-5 (1.51e-5) - | 1.1714e-4 (1.54e-5) + | 2.6775e-4 (2.15e-5) + | 1.0231e-4 (1.13e-5) |
|  | 15 | 24 | 2.3425e-7 (2.00e-8) + | 0.0000e+0 (0.00e+0) - | 8.207e-11 (1.33e-10) - | 1.175e-11 (2.74e-11) - | 5.2318e-8 (2.96e-8) - | 5.4635e-8 (2.07e-8) - | 6.7915e-8 (1.21e-7) |
| MaF4 | 5 | 14 | 7.0758e-3 (1.18e-2) - | 1.0363e-2 (1.61e-2) - | 3.9763e-3 (3.79e-3) - | 0.0000e+0 (0.00e+0) - | 3.0653e-3 (6.83e-3) - | 1.0704e-2 (2.20e-2) - | 2.6528e-2 (1.66e-2) |
|  | 10 | 19 | 2.2129e-5 (4.07e-5) = | 1.2256e-5 (1.96e-5) = | 1.5490e-7 (6.86e-7) = | 0.0000e+0 (0.00e+0) - | 2.1651e-9 (6.52e-9) - | 4.1800e-7 (2.29e-6) - | 4.4478e-7 (1.41e-6) |
|  | 15 | 24 | 3.3164e-10 (1.77e-9) = | 1.912e-10 (7.48e-10) = | 7.315e-12 (3.21e-11) + | 0.0000e+0 (0.00e+0) - | 8.047e-16 (3.09e-15) = | 1.100e-20 (6.03e-20) - | 1.1844e-12 (3.90e-12) |
| MaF7 | 5 | 24 | 2.3400e-1 (3.31e-3) + | 2.4239e-1 (4.20e-3) + | 8.9655e-2 (3.73e-2) - | 0.0000e+0 (0.00e+0) - | 2.0654e-1 (5.39e-3) + | 2.5981e-1 (5.95e-3) + | 1.3815e-1 (1.89e-2) |
|  | 10 | 29 | 9.2672e-2 (3.39e-2) + | 4.3467e-3 (6.72e-3) - | 6.6765e-3 (1.01e-2) - | 0.0000e+0 (0.00e+0) - | 7.4662e-2 (3.65e-2) + | 1.0171e-1 (9.53e-3) + | 4.5620e-2 (2.25e-2) |
|  | 15 | 34 | 1.3481e-1 (1.43e-2) + | 1.4728e-7 (8.07e-7) - | 4.1105e-3 (1.04e-2) - | 0.0000e+0 (0.00e+0) - | 7.6569e-2 (2.41e-2) + | 7.2532e-3 (4.75e-3) + | 2.8273e-3 (5.77e-3) |
| MaF9 | 5 | 2 | 1.6894e-1 (3.44e-2) - | 2.1244e-1 (5.50e-2) - | 2.3630e-1 (1.48e-2) - | 8.9734e-2 (8.06e-2) - | 1.7811e-1 (1.47e-2) - | 2.5512e-1 (2.99e-2) = | 2.5799e-1 (3.94e-2) |
|  | 10 | 2 | 3.7907e-3 (2.32e-3) - | 2.5043e-4 (4.99e-4) - | 2.5049e-3 (1.41e-3) - | 1.8771e-3 (2.52e-3) - | 4.1887e-3 (1.38e-3) - | 0.0000e+0 (0.00e+0) - | 1.3621e-2 (1.21e-3) |
|  | 15 | 2 | 3.4479e-4 (2.96e-4) - | 2.0722e-5 (5.01e-5) - | 1.0868e-4 (1.28e-4) - | 2.3179e-5 (8.41e-5) - | 1.1873e-4 (8.35e-5) - | 3.1195e-4 (3.65e-4) - | 7.5973e-4 (3.50e-4) |
| WFG1 | 5 | 14 | 8.2782e-1 (4.67e-2) - | 8.8549e-1 (2.97e-2) - | 9.8535e-1 (2.14e-2) + | 8.8677e-1 (5.77e-2) - | 8.2513e-1 (6.43e-2) - | 9.2670e-1 (2.91e-2) - | 9.8014e-1 (3.20e-2) |
|  | 10 | 19 | 8.1138e-1 (7.04e-2) - | 7.2826e-1 (9.16e-2) - | 9.3665e-1 (5.49e-2) - | 8.3991e-1 (1.52e-1) - | 7.7880e-1 (8.98e-2) - | 4.5727e-1 (4.77e-2) - | 9.8454e-1 (3.66e-2) |
|  | 15 | 24 | 8.4640e-1 (8.98e-2) - | 6.4749e-1 (1.25e-1) - | 9.9971e-1 (5.22e-4) + | 9.5269e-1 (8.83e-2) - | 8.7719e-1 (7.65e-2) - | 3.9253e-1 (5.10e-2) - | 9.8975e-1 (3.57e-2) |
| WFG8 | 5 | 14 | 5.5626e-1 (1.25e-2) - | 6.2138e-1 (6.58e-3) - | 5.5515e-1 (1.40e-2) - | 6.2413e-1 (5.66e-3) - | 6.2724e-1 (1.32e-2) - | 6.5656e-1 (4.99e-3) + | 6.3811e-1 (5.59e-3) |
|  | 10 | 19 | 7.8025e-1 (1.30e-2) - | 5.3455e-1 (3.71e-2) - | 6.4987e-1 (7.22e-2) - | 6.6853e-1 (1.10e-2) - | 6.0412e-1 (9.23e-2) - | 6.4226e-1 (4.04e-2) - | 7.9873e-1 (1.06e-2) |
|  | 15 | 24 | 8.0856e-1 (5.30e-2) = | 4.5034e-1 (3.39e-2) - | 7.5488e-1 (2.37e-2) - | 6.0761e-1 (1.60e-2) - | 5.3066e-1 (1.54e-1) - | 4.0342e-1 (4.59e-2) - | 8.3792e-1 (2.56e-2) |
| +/-/= |  |  | 9/18/3 | 3/24/3 | 6/19/5 | 0/30/0 | 10/19/1 | 9/20/1 |  |

Table 2. IGD values of solution sets obtained by seven algorithms on regular and irregular test instances

| Problem | M | D | ANSGAIII | AdaW | MOEADAWA | MOEADDU | RVEA* | NRVMOEA | IMOEA |
|---|---|---|---|---|---|---|---|---|---|
| DTLZ1 | 5 | 9 | 9.6786e-2 (4.59e-2) - | 8.5440e-2 (4.28e-2) - | 6.6602e-2 (1.97e-3) - | 7.3444e-2 (2.79e-2) = | 6.9195e-2 (1.65e-2) - | 8.9340e-2 (4.26e-2) - | 6.4955e-2 (1.36e-3) |
| | 10 | 14 | 4.0336e-1 (2.89e-1) - | 3.4574e+0 (2.85e+0) - | 1.4843e-1 (1.20e-2) = | 1.6449e-1 (7.10e-2) - | 1.4412e-1 (7.82e-2) - | 9.5689e+0 (3.29e+0) - | 1.4394e-1 (2.16e-2) |
| | 15 | 19 | 3.6043e-1 (2.19e-1) - | 0.0000e+0 (0.00e+0) - | 9.4201e-1 (2.83e-2) + | 4.5561e-1 (3.53e-1) - | 9.4902e-1 (1.00e-1) + | 0.0000e+0 (0.00e+0) - | 9.0572e-1 (1.66e-1) |
| DTLZ3 | 5 | 14 | 2.2242e-1 (5.05e-2) - | 2.0165e-1 (1.24e-3) - | 2.0187e-1 (1.67e-3) - | 1.9961e-1 (1.30e-3) - | 1.9571e-1 (8.26e-5) - | 2.0381e-1 (1.29e-3) - | 1.9100e-1 (1.10e-3) |
| | 10 | 19 | 5.4990e-1 (4.62e-2) - | 7.4361e-1 (7.50e-2) - | 5.0731e-1 (5.67e-2) - | 4.6412e-1 (2.12e-3) = | 4.3778e-1 (1.22e-3) + | 4.6628e-1 (5.97e-3) = | 4.7541e-1 (2.85e-2) |
| | 15 | 24 | 6.4267e-1 (1.46e-2) + | 1.5078e-2 (2.04e-2) - | 8.7101e-1 (7.73e-2) = | 8.6067e-1 (3.10e-2) = | 9.8546e-1 (1.37e-2) + | 8.6098e-1 (3.18e-2) = | 8.5383e-1 (3.63e-2) |
| DTLZ5 | 5 | 14 | 1.4880e-1 (3.30e-2) - | 1.2835e-1 (3.23e-2) - | 7.0961e-2 (9.21e-4) - | 1.6851e-1 (2.87e-2) - | 3.6415e-1 (9.15e-2) - | 8.1914e-2 (2.01e-2) - | 1.9312e-2 (3.38e-3) |
| | 10 | 19 | 3.3658e-1 (8.49e-2) - | 2.9534e-1 (8.86e-2) - | 8.2658e-2 (1.46e-2) - | 2.4781e-1 (3.52e-2) - | 4.4895e-1 (1.99e-1) - | 1.8995e-1 (3.10e-2) - | 2.0521e-2 (5.62e-3) |
| | 15 | 24 | 2.8010e-1 (5.27e-2) - | 2.6973e-2 (2.40e-2) - | 9.2253e-2 (6.24e-4) - | 9.0421e-2 (8.31e-4) - | 9.1098e-2 (3.89e-4) - | 2.7113e-2 (2.34e-2) - | 9.5145e-2 (2.19e-4) |
| DTLZ7 | 5 | 24 | 3.5431e-1 (3.19e-2) + | 3.0465e-1 (7.42e-3) + | 5.6228e-1 (3.12e-2) - | 1.1997e+1 (1.58e+0) - | 4.5921e-1 (1.96e-2) = | 2.8473e-1 (6.43e-2) + | 4.4109e-1 (3.30e-2) |
| | 10 | 29 | 2.1569e+0 (7.78e-1) - | 2.4706e+0 (9.55e-1) - | 1.3254e+0 (1.41e-1) - | 3.0237e+1 (2.85e+0) - | 1.5972e+0 (1.95e-1) - | 8.7008e-1 (1.52e-2) + | 1.2120e+0 (7.37e-2) |
| | 15 | 34 | 8.2886e+0 (6.61e-1) - | 4.3856e-8 (2.40e-7) - | 2.4160e-3 (6.72e-3) = | 0.0000e+0 (0.00e+0) - | 6.7935e-2 (2.35e-2) + | 1.2212e-2 (8.34e-3) + | 4.1368e-3 (6.71e-3) |
| IDTLZ2 | 5 | 14 | 2.7254e-1 (1.90e-2) - | 1.9391e-1 (1.68e-3) + | 2.4891e-1 (4.43e-3) - | 3.2345e-1 (1.07e-2) - | 3.1351e-1 (2.27e-2) - | 2.4477e-1 (1.18e-2) - | 2.2195e-1 (3.30e-3) |
| | 10 | 19 | 6.6535e-1 (1.30e-2) - | 4.5944e-1 (5.70e-3) - | 8.1368e-1 (3.59e-2) - | 7.5497e-1 (2.28e-2) - | 7.3538e-1 (1.51e-2) - | 4.9671e-1 (1.26e-2) - | 4.2743e-1 (3.44e-3) |
| | 15 | 24 | 7.5195e-1 (7.32e-3) - | 0.0000e+0 (0.00e+0) - | 8.207e-11 (1.33e-10) - | 1.175e-11 (2.74e-11) - | 5.2318e-8 (2.96e-8) - | 5.4635e-8 (2.07e-8) - | 6.7915e-8 (1.21e-7) |
| MaF4 | 5 | 14 | 2.1100e+1 (2.10e+1) - | 1.8079e+1 (1.77e+1) - | 8.3119e+0 (4.68e+0) - | 1.9528e+2 (8.87e+1) - | 2.3928e+1 (2.13e+1) - | 1.8198e+1 (1.90e+1) - | 4.5835e+0 (4.22e+0) |
| | 10 | 19 | 4.268e+2 (3.55e+2) = | 3.854e+2 (6.47e+2) = | 2.4459e+2 (6.86e+1) - | 1.5546e+4 (6.42e+3) - | 1.0569e+3 (1.52e+3) - | 2.0697e+3 (1.65e+3) - | 2.2060e+2 (1.46e+2) |
| | 15 | 24 | 2.7363e+4 (2.44e+4) - | 1.912e-10(7.48e-10) = | 7.315-12 (3.21e-11) + | 0.0000e+0 (0.00e+0) - | 8.0479e-11(3.09e-11)- | 1.100e+2 (6.03e+2) + | 1.1844e-12 (3.90e-12) |
| MaF7 | 5 | 24 | 3.4849e-1 (1.91e-2) + | 3.0396e-1 (7.79e-3) + | 5.7220e-1 (2.65e-2) - | 1.1982e+1 (1.69e+0) - | 4.5987e-1 (1.94e-2) + | 3.0151e-1 (9.43e-2) + | 5.3971e-1 (3.84e-2) |
| | 10 | 29 | 2.0845e+0 (6.87e-1) - | 2.2241e+0 (8.35e-1) - | 1.4030e+0 (2.70e-1) - | 3.0824e+1 (2.07e+0) - | 1.6719e+0 (1.93e-1) - | 8.6761e-1 (1.65e-2) + | 1.2106e+0 (8.04e-2) |
| | 15 | 34 | 8.2696e+0 (7.83e-1) - | 1.4728e-7 (8.07e-7) - | 4.1105e-3 (1.04e-2) = | 0.0000e+0 (0.00e+0) - | 7.6569e-2 (2.41e-2) + | 7.2532e-3 (4.75e-3) + | 2.8273e-3 (5.77e-3) |
| MaF9 | 5 | 2 | 5.2130e-1 (1.48e-1) - | 3.7032e-1 (1.82e-1) - | 2.6708e-1 (3.88e-2) - | 1.0057e+0 (5.73e-1) - | 4.1672e-1 (5.59e-2) - | 2.4321e-1 (8.80e-2) = | 2.3769e-1 (1.24e-1) |
| | 10 | 2 | 1.9244e+0 (3.71e-0) - | 1.2260e+1 (2.97e+1) - | 1.8304e+0 (1.11e+0) - | 3.4475e+0 (2.36e+0) - | 1.0876e+0 (2.30e-1) - | 1.6647e+1 (5.44e+0) - | 2.8557e-1 (1.17e-1) |
| | 15 | 2 | 4.2830e+0 (5.35e+0) - | 2.0722e-5 (5.01e-5) - | 1.0868e-4 (1.28e-4) + | 2.3179e-5 (8.41e-5) - | 1.1873e-4 (8.35e-5) - | 3.1195e-4 (3.65e-4) - | 7.5973e-4 (3.50e-4) |
| WFG1 | 5 | 14 | 6.4824e-1 (6.53e-2) + | 6.0296e-1 (3.27e-2) + | 6.6857e-1 (4.98e-2) + | 6.8340e-1 (1.14e-1) + | 6.3276e-1 (1.20e-1) + | 4.8680e-1 (2.86e-2) + | 7.0721e-1 (4.48e-2) |
| | 10 | 19 | 1.5604e+0 (7.61e-2) + | 1.5281e+0 (1.36e-1) + | 1.5756e+0 (1.12e-1) + | 1.4904e+0 (2.88e-1) + | 1.3708e+0 (8.27e-2) + | 2.1529e+0 (1.72e-1) - | 1.7757e+0 (8.13e-2) |
| | 15 | 24 | 2.1679e+0 (8.99e-2) + | 6.4749e-1 (1.25e-1) - | 9.9971e-1 (5.22e-4) - | 9.5269e-1 (8.83e-2) - | 8.7719e-1 (7.65e-2) - | 3.9253e-1 (5.10e-2) - | 9.8975e-1 (3.57e-2) |
| WFG8 | 5 | 14 | 1.3134e+0 (2.10e+1) - | 1.1962e+0 (1.10e-2) + | 1.4259e+0 (2.60e-2) - | 1.3624e+0 (1.28e-2) - | 1.1915e+0 (7.69e-3) + | 1.1913e+0 (1.16e-2) + | 1.3060e+0 (3.26e-2) |
| | 10 | 19 | 5.3145e+0 (4.33e-1) - | 4.9808e+0 (8.32e-2) = | 5.4655e+0 (3.74e-1) - | 5.8603e+0 (1.48e-1) - | 5.4902e+0 (1.17e-1) - | 5.3041e+0 (1.01e-1) - | 5.0293e+0 (1.71e-1) |
| | 15 | 24 | 9.0363e+0 (3.43e-1) + | 9.0333e+0 (2.33e-1) + | 1.1732e+1 (3.88e-1) - | 1.2391e+0 (1.48e-1) - | 9.9113e+0 (5.80e-1) + | 9.9148e+0 (2.12e-1) + | 1.0754e+1(5.78e-1) |
| +/-/= | | | 7/22/1 | 6/21/3 | 5/21/4 | 2/24/4 | 10/19/1 | 9/18/3 | |

Here +/-/= indicates whether the compared algorithms are significantly better, worse, or equivalent to IMOEA, the best result on each test instance is highlighted in gray.

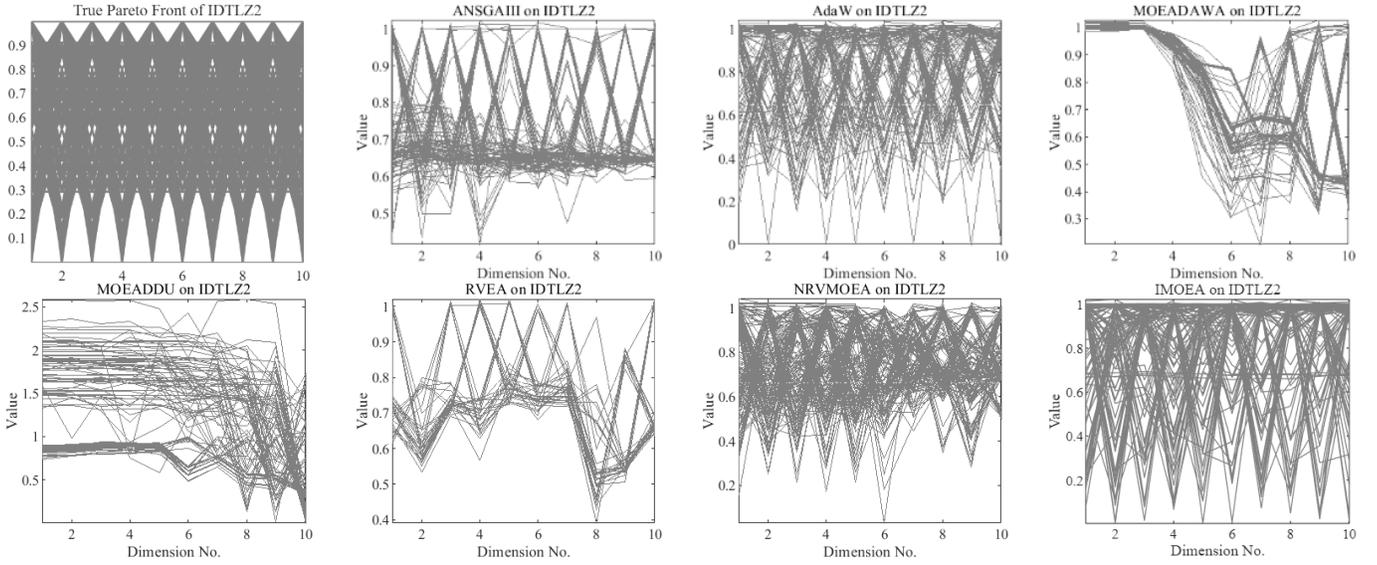

Fig.2.Nondominated solution sets obtained by ANSGA-III, Adaw, MOEA/D-AWA, MOEA/DU, RVEA, NRVMOEA and IMOEA on solving IDTLZ2 with ten objectives.

## A. Parameter Settings

The proposed IMOEA and all the comparison algorithms adopt SBX [26] as the crossover operator, with the crossover probability $P_c = 1.0$ and the crossover distribution index $n_c = 20$; polynomial mutation [27] is used as the mutation operator, with the mutation probability $P_m = 1/D$ (D is the dimension of the decision space) and the mutation distribution index $n_m = 20$. According to the different objective dimensions, the experiments on the above-mentioned test sets were conducted under objective dimensions of 5, 10, and 15 respectively. Correspondingly, the population size was set to 120, 169, and 220 respectively, and the maximum evaluation times were set to 30,000, 30,000, and 60,000 respectively. For each test case, all comparison algorithms were independently run 30 times.

## B. Performance indicators

This study adopts Invert Generational Distance (IGD) as the performance indicator. The calculation of IGD is as follows:

$$IGD(P, P^*) = \frac{1}{|P^*|} \sum_{v \epsilon P^*} d(v, P) \qquad (11)$$

Where, $d(v, P)$ represents the Euclidean distance from point $v$ in $P^*$ to the nearest point in P, and denotes the cardinality of $|P^*|$.

This study adopts another comprehensive performance indicator hypervolume (HV), and its calculation is as follows:

$$HV = L\left(\cup_{i=1}^{|EA|} v_i\right) \qquad (12)$$

Where, $L$ represents the Lebesgue measure, $v_i$ denotes the hypervolume formed by the reference point and the non-dominated individuals, and $EA$ represents the non-dominated sets.

## C. Effectiveness of threshold

The threshold is related to the rate of $R_2^{new}$. In Fig.3., we conduct experiments on four irregular test instances. In the early and middle stages, the population in the exploration phase, so a loose threshold is used to facilitate a more effective search, that is $TH = 0.1 * (1 + (m - 5))(m \geq 5)$; In the middle and later stages, which are mainly the development phase, a relatively tight threshold is set to promote population development, $TH = 0.05 * (1 + (M - 5))$.

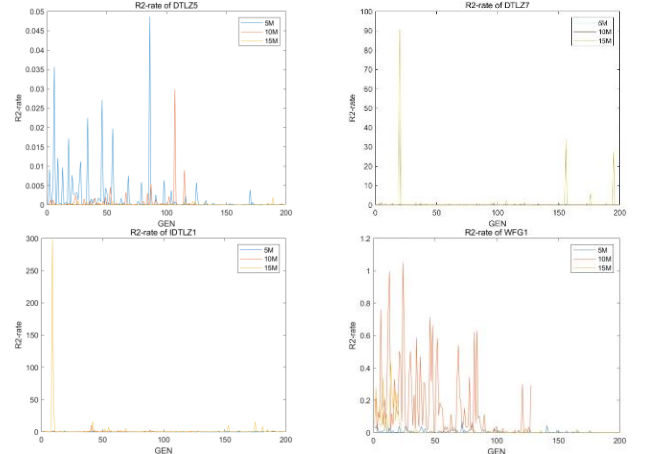

Fig.3. The trend graph of $Improvement_{rate}$ on DTLZ5, DTLZ7, IDTLZ1 and WFG1 with the change of generations

## D. Experimental Result

In Fig.2. shows nondominated solution sets obtained by ANSGA-III, Adaw, MOEA/D-AWA, MOEA/DU, RVEA, NRVMOEA and IMOEA on solving IDTLZ2 with ten objectives, it can be seen from these results that the approximate PFs obtained through IMOEA are very close to the true PFs.

In Table 1, the HV values of each algorithm on 5, 10, and 15-dimensional objective spaces are listed. From the results,

ANSGA-III achieves the best results on 2 test instances, MOEA/D-AWA on 3 test instances, RVEA* on 2 test instances, and NRVMOEA on 7 test instances. In terms of algorithm comparison, ANSGA_III, Adaw, MOEA/D-AWA, MOEA/D-DU, RVEA*, and NRVMOEA outperform the algorithm proposed in this study on 9, 3, 6, 0, 10, and 9 test instances respectively, underperform on 18, 24, 19, 30, 19, and 20 test instances respectively, and are similar on 3, 3, 5, 0, 1, and 1 test instances respectively.

In Table 2, the IGD values of each algorithm on 5, 10, and 15-dimensional objective spaces are listed. From the results, ANSGA-III achieves the best results on 2 test instances, Adaw on 2 test instances, MOEA/D-AWA on 1 test instances, RVEA* on 2 test instances, and NRVMOEA on 7 test instances. In terms of algorithm comparison, ANSGA_III, Adaw, MOEA/D-AWA, MOEA/D-DU, RVEA*, and NRVMOEA outperform the algorithm proposed in this study on 7, 6, 5, 2, 10, and 8 test instances respectively, underperform on 21, 21, 21, 24, 19, and 18 test instances respectively, and are similar on 1, 3, 4, 4, 1, and 3 test instances respectively.

## V. Conclusions

This study proposes a method that integrates indicator techniques with decomposition methods, achieving adaptive dynamic adjustment of the weight vector. Specifically, IMOEA is based on the MOEA/D framework and uses a simplified hypervolume indicator to determine the sparsity among individuals, thereby deleting or adding sub-problems in critical regions. And it employs the $R_2^{new}$ to dynamically control the update frequency of the weight vector, which can effectively guide the evolution of populations, providing an efficient and highly accurate solution for many-objective optimization problems.

Future research may investigate more computationally efficient hypervolume-based methods that effectively preserve accuracy.

## References

[1] V. Palakonda and J.-M. Kang, "Many-objective real-world engineering problems: a comparative study of state-of-the-art algorithms," *IEEE Access*, vol. 11, pp. 111636–111654, 2023.

[2] C. Felipe Coello Castillo and C. A. Coello, "A survey of applications of multi-objective evolutionary algorithms in biotechnology," in *2024 IEEE Congress on Evolutionary Computation (CEC)*, June 2024, pp. 1–8.

[3] X. Cai, H. Sun, Q. Zhang, and Y. Huang, "A grid weighted sum pareto local search for combinatorial multi and many-objective optimization," *IEEE Trans. Cybern.*, vol. 49, no. 9, pp. 3586–3598, Sept. 2019.

[4] S. Wang et al., "A pareto dominance relation based on reference vectors for evolutionary many-objective optimization," *Appl. Soft Comput.*, vol. 157, pp. 111505, May 2024.

[5] G. Lopes, K. Klamroth, and L. Paquete, "A greedy hypervolume polychotomic scheme for multiobjective combinatorial optimization," *Comput. Oper. Res.*, vol. 183, pp. 107140, Nov. 2025.

[6] Z. Wang, C. Xiao, and A. Zhou, "Exact calculation of inverted generational distance," *IEEE Trans. Evol. Comput.*, pp. 1–1, 2024.

[7] A. Jaszkiewicz and P. Zielniewicz, "Exact calculation and properties of the R2 multiobjective quality indicator," *IEEE Trans. Evol. Comput.*, vol. 29, no. 4, pp. 1227–1238, Aug. 2025.

[8] H. Ishibuchi, Y. Setoguchi, H. Masuda, and Y. Nojima, "Performance of decomposition-based many-objective algorithms strongly depends on pareto front shapes," *IEEE Trans. Evol. Comput.*, vol. 21, no. 2, pp. 169–190, Apr. 2017.

[9] Q. Zhang and H. Li, "MOEA/D: a multiobjective evolutionary algorithm based on decomposition," *IEEE Trans. Evol. Comput.*, vol. 11, no. 6, pp. 712–731, Dec. 2007.

[10] I. Das and J. E. Dennis, "Normal-boundary intersection: a new method for generating the pareto surface in nonlinear multicriteria optimization problems," *SIAM J. Optim.*, vol. 8, no. 3, pp. 631–657, Aug. 1998.

[11] K. Li, K. Deb, Q. Zhang, and S. Kwong, "An evolutionary many-objective optimization algorithm based on dominance and decomposition," *IEEE Trans. Evol. Comput.*, vol. 19, no. 5, pp. 694–716, Oct. 2015.

[12] A. Jaszkiewicz, "On the performance of multiple-objective genetic local search on the 0/1 knapsack problem - a comparative experiment," *IEEE Trans. Evol. Comput.*, vol. 6, no. 4, pp. 402–412, Aug. 2002.

[13] M. Li and X. Yao, "What weights work for you? Adapting weights for any pareto front shape in decomposition-based evolutionary multi-objective optimisation," Sept. 2017.

[14] Y. Qi, X. Ma, F. Liu, L. Jiao, J. Sun, and J. Wu, "MOEA/D with adaptive weight adjustment," *Evol. Comput.*, vol. 22, no. 2, pp. 231–264, June 2014.

[15] K. Shang, H. Ishibuchi, M.-L. Zhang, and Y. Liu, "A new R2 indicator for better hypervolume approximation," in *Proceedings of the Genetic and Evolutionary Computation Conference*, Kyoto Japan: ACM, pp. 745–752, July 2018.

[16] X. Ma, Q. Zhang, G. Tian, J. Yang, and Z. Zhu, "On tchebycheff decomposition approaches for multiobjective evolutionary optimization," *IEEE Trans. Evol. Comput.*, vol. 22, no. 2, pp. 226–244, Apr. 2018.

[17] H. Jain and K. Deb, "An evolutionary many-objective optimization algorithm using reference-point based nondominated sorting approach, part II: handling constraints and extending to an adaptive approach," *IEEE Trans. Evol. Comput.*, vol. 18, no. 4, pp. 602–622, Aug. 2014.

[18] R. Cheng, Y. Jin, M. Olhofer, and B. Sendhoff, "A reference vector guided evolutionary algorithm for many-objective optimization," *IEEE Trans. Evol. Comput.*, vol. 20, no. 5, pp. 773–791, Oct. 2016.

[19] Y. Yuan, H. Xu, B. Wang, B. Zhang, and X. Yao, "Balancing convergence and diversity in decomposition-based many-objective optimizers," *IEEE Trans. Evol. Comput.*, vol. 20, no. 2, pp. 180–198, Apr. 2016.

[20] Y. Hua, Q. Liu, and K. Hao, "Adaptive normal vector guided evolutionary multi- and many-objective optimization," *Complex Intell. Syst.*, vol. 10, no. 3, pp. 3709–3726, June 2024.

[21] K. Deb, L. Thiele, M. Laumanns, and E. Zitzler, "Scalable multi-objective optimization test problems," in *Proceedings of the 2002 Congress on Evolutionary Computation. CEC'02 (cat. No.02th8600)*, vol.1, pp. 825–830, May 2002.

[22] H. Jain and K. Deb, "An Improved Adaptive Approach for Elitist Nondominated Sorting Genetic Algorithm for Many-Objective Optimization," in *Evolutionary Multi-Criterion Optimization*, R. C. Purshouse, P. J. Fleming, C. M. Fonseca, S. Greco, and J. Shaw, Eds., Berlin, Heidelberg: Springer, pp. 307–321 2013.

[23] S. Huband, P. Hingston, L. Barone, and L. While, "A review of multiobjective test problems and a scalable test problem toolkit," *IEEE Transactions on Evolutionary Computation*, vol. 10, no. 5, pp. 477–506, Oct. 2006.

[24] R. Cheng et al., "A benchmark test suite for evolutionary many-objective optimization," *Complex Intell. Syst.*, vol. 3, no. 1, pp. 67–81, Mar. 2017.

[25] Y. Tian, R. Cheng, X. Zhang, and Y. Jin, "PlatEMO: A MATLAB Platform for Evolutionary Multi-Objective Optimization," *IEEE Computational Intelligence Magazine*, vol. 12, no. 4, pp. 73–87, Nov. 2017.

[26] R. B. Agrawal, K. Deb, and R. B. Agrawal, "Simulated binary crossover for continuous search space," *Complex Systems*, vol. 9, no. 3, pp. 115–148, 1994.

[27] K. Deb and M. Goyal, "A combined genetic adaptive search (GeneAS) for engineering design," *Computer Science and Informatics*, vol. 26, no. 4, 1996.